\documentclass[letterpaper, 10pt, conference]{ieeeconf} 
\IEEEoverridecommandlockouts
\usepackage{cite}
\usepackage{amsmath,amssymb,amsfonts}
\usepackage{algorithmic}
\usepackage{graphicx}
\usepackage{capt-of}
\usepackage{textcomp}
\usepackage{xcolor}
\usepackage{booktabs}
\usepackage{colortbl}
\usepackage{fontawesome5}
\usepackage[hidelinks]{hyperref}
\hypersetup{
    colorlinks=true,
    linkcolor=blue,      
    citecolor=red,       
    urlcolor=cyan
}

\def\BibTeX{{\rm B\kern-.05em{\sc i\kern-.025em b}\kern-.08em
    T\kern-.1667em\lower.7ex\hbox{E}\kern-.125emX}}
\begin{document}

\title{TacVLA: Contact-Aware \textbf{Tac}tile Fusion \\for Robust \textbf{V}ision-\textbf{L}anguage-\textbf{A}ction Manipulation
}

\author{ 
Kaidi Zhang$^{*, 3}$, Heng Zhang$^{*, 1, 2, 3}$, Zhengtong Xu$^{3}$, Zhiyuan Zhang$^{3}$, Md Rakibul Islam Prince$^{4}$, \\
Xiang Li$^{3}$, Xiaojing Han$^{3}$, Yuhao Zhou$^{3}$, Arash Ajoudani$^{1}$, Yu She$^{3}$
\thanks{$^{*}$These authors contributed equally.}%
\thanks{$^{1}$Human-Robot Interfaces and Interaction Lab, Istituto Italiano di Tecnologia, Genova, Italy.}%
\thanks{$^{2}$Ph.D. program of national interest in Robotics and Intelligent Machines (DRIM) and Universit\`a di Genova, Genoa, Italy.}%
\thanks{$^{3}$Kaidi, Heng, Zhengtong, Zhiyuan, Prince, Xiang, Xiaojing, Yuhao, and Yu are with the Edwardson School of Industrial Engineering, Purdue University, West Lafayette, IN 47907, USA \{zhan5896, zhan6025, xu1703, zhan5570, prince26, li5315, han708, zhan5570, zhou1437, shey\}@purdue.edu. }%
\thanks{$^{4}$Elmore Family School of Electrical and Computer Engineering, Purdue University, West Lafayette, IN 47907, USA, prince26@purdue.edu. }%
}

\IEEEaftertitletext{%
\begin{center}
\includegraphics[width=0.95\textwidth]{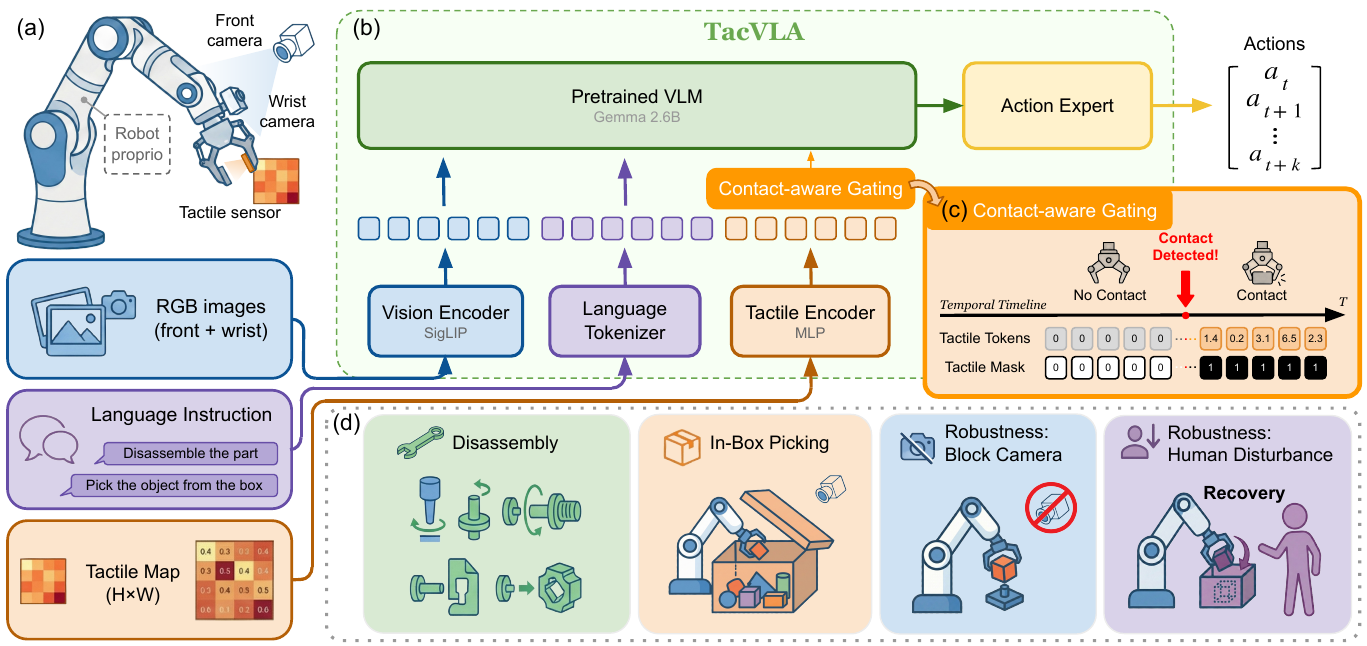}
\captionof{figure}{\textbf{Overview of TacVLA}. 
(a) Input modalities including visual observations, language instructions, and tactile measurements. 
(b) TacVLA architecture, consisting of modality-specific encoders and tokenizer, a pretrained VLM backbone, an action expert, and the contact-aware gating module.
(c) The proposed contact-aware gating module that selectively activates tactile tokens based on the contact state, enabling adaptive multimodal fusion during contact-rich manipulation. 
(d) Experimental evaluation on contact-rich constraint-locked disassembly and in-box picking tasks, together with robustness tests under camera occlusion and human disturbance.}
\label{fig:architecture}
\end{center}
}

\maketitle

\begin{abstract}
Vision-Language-Action (VLA) models have demonstrated significant advantages in robotic manipulation. 
However, their reliance on vision and language often leads to suboptimal performance in tasks involving visual occlusion, fine-grained manipulation, and physical contact.
To address these challenges, we propose TacVLA, a fine-tuned VLA model by incorporating tactile modalities into the transformer-based policy to enhance fine-grained manipulation capabilities. 
Specifically, we introduce a contact-aware gating mechanism that selectively activates tactile tokens only when contact is detected, enabling adaptive multimodal fusion while avoiding irrelevant tactile interference.
The fused visual, language, and tactile tokens are jointly processed within the transformer architecture to strengthen cross-modal grounding during contact-rich interaction.
Extensive experiments on constraint-locked disassembly, in-box picking and robustness evaluations demonstrate that TacVLA outperforms baselines,
improving the performance by averaging 20\% success rate in disassembly and 60\% in in-box picking, achieving a 2.1$\times$ improvement under visual occlusion, and showing recovery behavior under human disturbance.
Videos are available at https://sites.google.com/view/tacvla.
\end{abstract}

\section{Introduction}
Vision-Language-Action (VLA) models have demonstrated remarkable potential in achieving generalized robotic manipulation \cite{pi0.5,kim2024openvla,brohan2023rt2}. 
By integrating pretrained Vision Language Models (VLMs), these models can effectively interpret complex natural language instructions and perceive unstructured environments. 
Furthermore, their ability to reason casually and plan sequentially allows them to bridge the gap between high-level semantic understanding and low-level control policies~\cite{pi0.5,kim2024openvla,brohan2023rt2,zhang2025safe}. 
Consequently, they provide strong general-purpose backbones that support deployment across diverse robot embodiments and real-world scenarios.

Despite these advancements, a significant limitation of current VLA architectures is their predominant reliance on vision and language as the sole modalities for perceiving the physical world~\cite{bianchini2025vysics}. 
This visual dependency becomes a critical bottleneck in tasks involving severe visual occlusions, where the robot's end-effector or the manipulated object itself obstructs the camera's view. 
Moreover, for fine-grained contact-rich manipulation tasks such as disassembly or precise assembly, visual signals alone often fail to capture essential physical properties like contact forces, friction, and surface textures~\cite{zhang2025safe, luu2025manifeel}. 
Without direct feedback from physical contact, robots struggle to dynamically adjust their actions in response to unexpected resistance or slippage, leading to suboptimal performance~\cite{tsuji2025survey,xue2025slowfast,yu2025mimictouch}. 
Therefore, integrating complementary modalities that can sense physical interactions is essential for enhancing the robustness and precision of VLA models in contact-rich environments.

Recent studies have begun incorporating tactile sensing into VLA-style policies~\cite{hao2025tla,zhang2025vtla,yu2025octopi,cheng2025omnivtla,bi2025vlatouch,liu2025mla,huang2025tactile}, showing that touch complements vision and language when visual information is limited or occluded.
Tactile feedback provides direct signals of contact events, slippage, and local geometry, enabling more reliable grasp stabilization and corrective adjustments in insertion~\cite{hao2025tla,zhang2025vtla}, wiping~\cite{huang2025tactile,bi2025vlatouch}, and fruit peeling~\cite{bi2025vlatouch} tasks.
However, most existing evaluations have limited emphasis on contact-intensive fine-grained manipulation tasks that require sustained and precise contact control, such as disassembly. 
In addition, the robustness of these systems under challenging conditions, such as visual occlusion, human perturbations, and dynamic contact variations, remains underexplored.
Moreover, many existing approaches treat tactile observations as image-like inputs~\cite{hao2025tla,zhang2025vtla,cheng2025omnivtla,bi2025vlatouch}, relying on dense pixel representations that increase token length and computational cost in transformer architectures.
Simply concatenating tactile tokens with visual and language tokens can also lead to inefficient cross-modal interaction, particularly during non-contact phases.
This raises a key question: \textbf{how should tactile modality be effectively integrated into VLA architectures to support robust fine-grained manipulation?}

To address these challenges, we propose TacVLA, an extension of VLA that integrates a compact tactile array into the transformer-based policy for contact-rich manipulation.
Instead of treating tactile signals as image-like inputs, we represent them as low-dimensional tactile tokens, enabling efficient multimodal processing without substantially increasing token length.
We further introduce a contact-aware gating mechanism that selectively activates tactile tokens only when physical contact is detected.
This design allows the model to leverage tactile feedback during interaction while avoiding unnecessary cross-modal interference during non-contact phases.
Through this adaptive fusion strategy, TacVLA strengthens grounding across vision, language, and touch while maintaining computational efficiency.

In summary, our contributions are threefold:
\begin{itemize}
    \item We introduce TacVLA, a tactile-enhanced extension of VLA that enables efficient multimodal integration for contact-rich fine-grained manipulation.
    
    \item We propose a compact tactile tokenization scheme and a contact-aware gating mechanism that adaptively regulates cross-modal interaction within the transformer architecture.
    
    \item We validate our approach on contact-rich constraint-locked disassembly, in-box picking and robustness benchmarks, demonstrating consistent improvements 
    particularly under visual occlusion and dynamic physical interactions.
\end{itemize}

\begin{figure}
    \centerline{\includegraphics[width=0.7\linewidth]{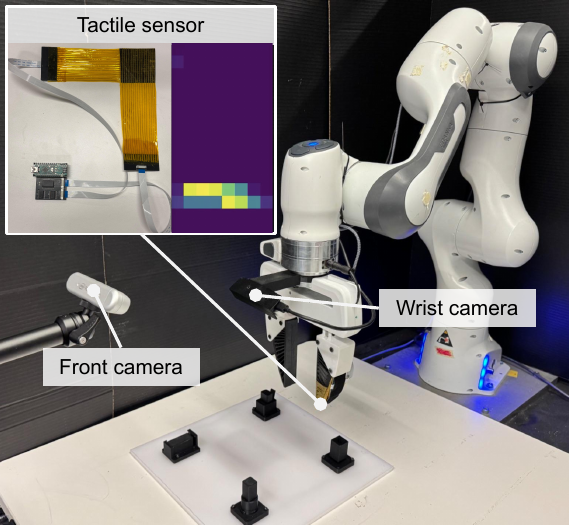}}
    \caption{Hardware setup: we utilize a 7 DoF Franka robotic platform equipped with tactile sensors and two cameras for visual input to evaluate our TacVLA model on contact-rich manipulation tasks.}
    \label{fig:hardware}
\end{figure}

\begin{figure}
    \centerline{\includegraphics[width=0.85\linewidth]{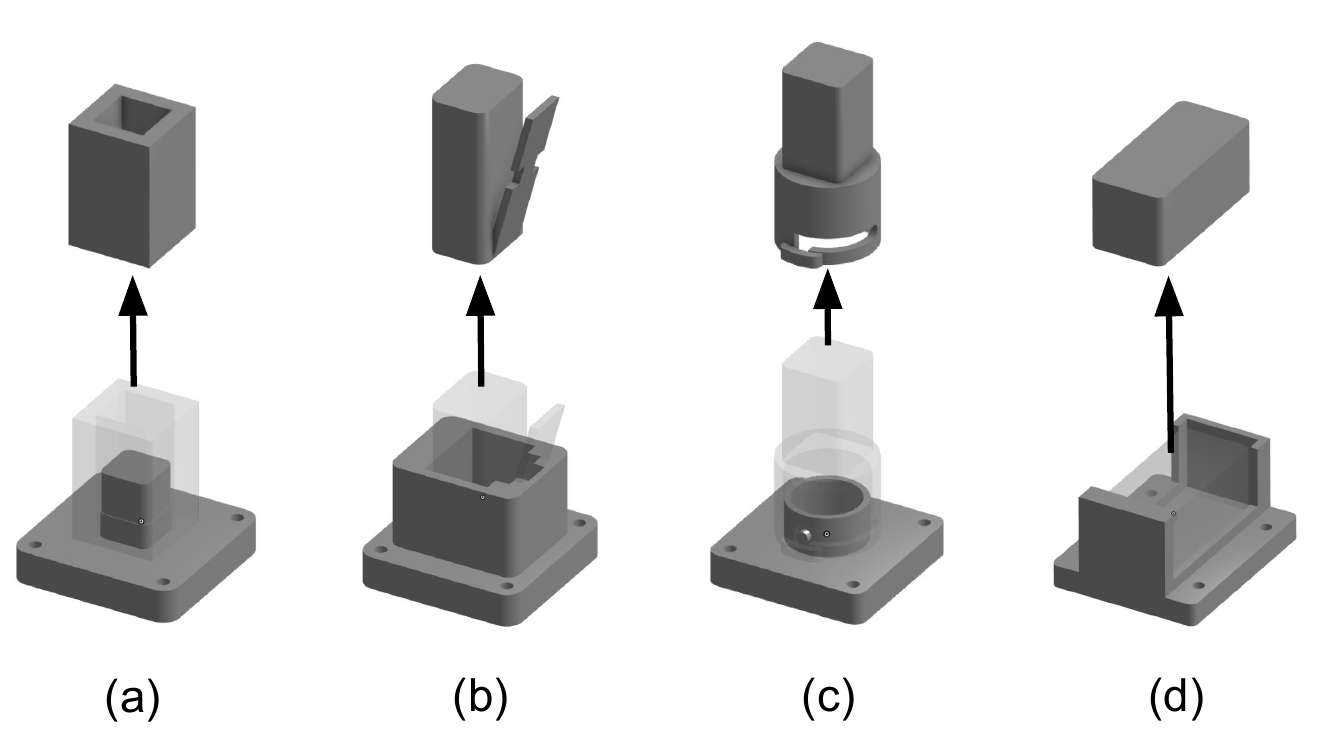}}
    \vspace{-0.8em}
    \caption{Four contact-rich constraint-locked disassembly tasks with diverse geometric constraints: (a) Task 1: tight shaft; (b) Task 2: press clip; (c) Task 3: shaft rotation; (d) Task 4: slide pull.}
    \label{fig:disassembly}
\end{figure}

\begin{figure*}
    \centering
    \includegraphics[width=0.68\textwidth]{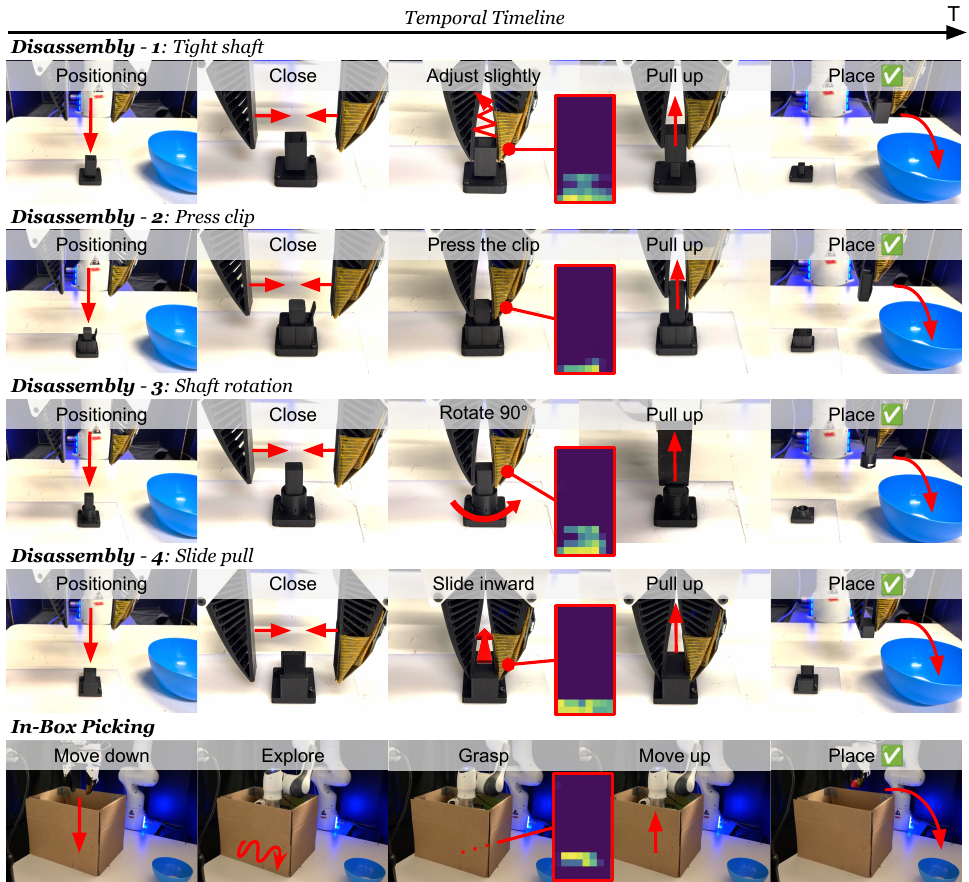}
    \caption{The real-world experimental setup and procedures for four constraint-locked disassembly and in-box picking task. The experiments demonstrate the capability of TacVLA in contact-rich fine-grained manipulation, as well as its robustness to visual occlusion in the in-box picking scenario.}
    \label{fig:real_world_setup}
\end{figure*}

\section{Related Work}

\subsection{VLA for Physical Interaction}\label{sec:vla_physical}
VLA models have emerged as a powerful framework for enabling robots to understand and execute complex tasks by integrating visual perception, natural language understanding, and action generation~\cite{kim2024openvla,pi0.5,brohan2023rt2,zhang2026SurveyPhysicalAI}.
By leveraging large-scale pretrained models, VLA systems can generalize across a wide range of tasks and environments, making them particularly effective for physical interaction scenarios. 
The ability of VLA models to reason about the relationships between objects, actions, and language allows them to perform tasks that require a deep understanding of the environment, such as object manipulation~\cite{zhang2026compliantvla, yu2025forcevla} and human-robot interaction~\cite{yin2025mivla}. 
However, the reliance on visual and linguistic inputs alone can limit their performance in scenarios where visual information is occluded~\cite{bianchini2025vysics, zhang2025canonical,equiform}, or when fine-grained physical interactions are required~\cite{zhang2025safe}, highlighting the need for incorporating additional sensory modalities.

In contrast, our work focuses on enhancing VLA models by integrating tactile sensing, which provides complementary information about physical interactions that are not easily captured through vision alone. 

\subsection{Tactile Sensing for Contact-rich Manipulation}\label{sec:tactile_manip} 
Tactile sensing provides rich, contact-centric feedback that complements vision for contact-rich manipulation tasks. It directly measures contact mechanics, normal and shear forces~\cite{wang2021gelsightwedge, zhang2023gelflow}, pressure distribution, local surface geometry~\cite{yuan2017gelsight,wang2021gelsightwedge, zhang2024gelroller}, incipient slip, and even acoustic feedback~\cite{zhang2025vibecheck}. 
Prior work has shown that leveraging tactile feedback improves reliability in contact-rich tasks by detecting subtle pose and alignment errors  as well as contact transitions~\cite{zhang2025vibecheck,xu2025unit}, which are crucial for robustness and safety in physical interaction.

Vision-based tactile sensors such as GelSight~\cite{yuan2017gelsight} provide high-resolution contact observations but introduce additional sensing complexity.
Recent visuotactile manipulation policies demonstrate that incorporating tactile feedback improves robustness under visual uncertainty and during fine-grained contact phases~\cite{luu2025manifeel,xue2025slowfast,yu2025mimictouch}. These works highlight the importance of touch in enhancing physical grounding of visuomotor policies.

In this work, we adopt a compact tactile array that provides efficient contact-state information suitable for integration into vision-language-action models.



\begin{table*}[!t]
\centering
\caption{Prompt Description for Constraint-locked Disassembly and In-box Picking Task}
\renewcommand{\arraystretch}{1.2}
\setlength{\tabcolsep}{6pt}
\begin{tabular}{>{\centering\arraybackslash}p{0.22\textwidth} 
                >{\centering\arraybackslash}p{0.63\textwidth}}
\hline
\textbf{Task} & \textbf{Prompt} \\
\hline
Disassembly - Task 1 &
Disassemble the object from the shaft and place it in the bowl. \\
Disassembly - Task 2 &
Press the clip and pull the object out, then place it in the bowl. \\
Disassembly - Task 3 &
Twist the object 90 degrees and pull it out from the shaft, then place it in the bowl. \\
Disassembly - Task 4 &
Slide the object inward and pull it out, then place it in the bowl. \\
In-box Picking Task &
Explore inside the box until contact is established, then grasp the object and place it in the bowl. \\
\hline
\end{tabular}
\label{tab:prompt_description}
\end{table*}

\subsection{Multimodal Fusion for Vision Language Tactile Models}\label{sec:multimodal_fusion}



Beyond tactile sensing itself, an equally critical challenge lies in effective multimodal fusion. Tactile signals are inherently contact-dependent and locally informative, differing fundamentally from global visual observations and discrete language tokens~\cite{din2025multimodal}. This modality heterogeneity makes naive feature concatenation suboptimal, particularly in transformer-based architectures where token competition and modality imbalance may dilute informative signals.

Recent works have explored structured fusion strategies for visuotactile policies, including force-aware alignment, token-level integration, and attention-based cross-modal modeling~\cite{pertsch2025fast,liu2025mla,cheng2025omnivtla,hao2025tla}. While these approaches improve multimodal interaction, most adopt static fusion mechanisms that assume continuous participation of tactile inputs throughout the trajectory ~\cite{hao2025tla,cheng2025omnivtla,zhang2025vtla,bi2025vlatouch}.

In contrast, our work explicitly addresses the state-dependent nature of tactile information by introducing token-level modality arbitration conditioned on contact state, enabling selective activation of tactile representations during physically informative phases.

\section{Methodology}
In this section, we present the architecture and training procedure of our proposed TacVLA model.
The core idea behind TacVLA is to integrate tactile sensing into a transformer-based VLA framework, enabling efficient multimodal fusion and adaptive grounding for contact-rich manipulation tasks.
To achieve this, we design a unified tokenization scheme, and introduce a contact-aware gating mechanism that selectively activates tactile tokens based on contact state. See Fig.~\ref{fig:architecture} for an overview of the model architecture.

\subsection{Overall Architecture}
Our TacVLA architecture follows a transformer-based VLA framework, extended with tactile perception for contact-rich manipulation. 
As shown in Fig.~\ref{fig:architecture}(b), the model consists of four components: modality tokenizers, a pretrained VLM backbone, and an action expert, and a contact-aware gating module.

Visual observations from both the front and wrist-mounted cameras are encoded using a SigLIP~\cite{zhai2023siglip} visual encoder, producing a sequence of visual tokens per image.
Language instructions along with robot proprioception are tokenized using the PaliGemma tokenizer.
For tactile sensing, the $15 \times 8$ tactile array (Fig.~\ref{fig:hardware}) is embedded using a lightweight MLP-based encoder, which projects the tactile map into 36 tactile tokens. We further add fixed 2D sine-cosine positional embeddings to preserve spatial structure. This design yields a compact yet expressive representation that captures local contact patterns, global pressure distribution, and contact geometry.

Let the multimodal token sequence at time $t$ be
\[
\tilde{\mathbf{z}}_t =
[\mathbf{z}_t^{vis}, \mathbf{z}_t^{lan+pro}, \tilde{\mathbf{z}}_t^{tac}],
\]
where $\mathbf{z}_t^{vis}, \mathbf{z}_t^{lan+pro}$ denote visual tokens, language plus proprioceptive tokens, and $\tilde{\mathbf{z}}_t^{tac}$ denotes tactile tokens processed by the contact-aware gating module.
All modality tokens are concatenated into a shared token sequence and fed into a pretrained VLM backbone. 
A non-causal attention mechanism over this prefix allows vision, language, and tactile tokens to freely cross-attend, producing a deeply integrated contextual representation for downstream control.

The fused representation is then provided as a prefix to an action expert module, following the Pi0.5~\cite{pi0.5} design, which is trained with a flow-matching objective and predicts continuous action sequences. 
This policy predicts an action sequence conditioned on the multimodal tokens,
\[
\mathbf{a}_{t:t+H} \sim \pi_\theta(\mathbf{a}_{t:t+H} \mid \tilde{\mathbf{z}}_t).
\]

\subsection{Contact-aware Gating Tactile Fusion}
Naively concatenating tactile tokens throughout the trajectory can be suboptimal, as tactile signals are typically uninformative during non-contact phases and may introduce unnecessary attention tokens.

To address this, we introduce a contact-aware token gating module as demonstrated in Fig.~\ref{fig:architecture}(c). Physical contact is detected using a threshold-based criterion: contact is declared when the number of taxels exceeding a predefined pressure threshold surpasses a fixed count.
Rather than dynamically inserting or removing tokens, we maintain a fixed token structure and treat tactile tokens as conditional inputs. 



Let $c_t \in \{0,1\}$ denote the contact flag at time $t$, $c_t=1$ only when physical contact is detected. 
We apply a contact-dependent attention mask $M_t^{tac}$ and embedding gating to tactile tokens $\mathbf{z}_t^{tac}$,
\[
M_t^{tac} = c_t \cdot \mathbf{1},
\]
\[
\tilde{\mathbf{z}}_t^{tac} = c_t \cdot \mathbf{z}_t^{tac},
\]
where $M_t^{tac}$ denotes the tactile-related submatrix in the full attention mask, and $\mathbf{1}$ is an all-one matrix with the same size as this submatrix.
When $c_t = 0$, tactile tokens are excluded from the attention computation and cannot participate in cross-modal interaction. In addition, zeroing tactile tokens suppresses offsets from embedding layers or positional encodings and reduces the influence of pre-contact sensor noise or sensor drift.

By preserving a fixed token topology while enabling state-dependent modality routing, the proposed module allows tactile information to influence policy decisions only when physically informative.

\begin{table*}[!t]
\centering
\caption{Performance comparison across Constraint-locked Disassembly and In-Box Picking tasks}
\renewcommand{\arraystretch}{1.5}
\setlength{\tabcolsep}{6pt}
\begin{tabular}{lcccccc}
\toprule
\multicolumn{1}{c}{\textbf{Method}} & \multicolumn{5}{c}{\textbf{Constraint-locked Disassembly}} & \textbf{In-Box Picking} \\
\cmidrule(lr){2-6}
 & \textbf{Task 1} & \textbf{Task 2} & \textbf{Task 3} & \textbf{Task 4} & \textbf{Average} &  \\
\midrule
3D Diffusion Policy + Tactile & 8/20 (40\%) & 8/20 (40\%) & 9/20 (45\%) & 0/20 (0\%) & 31.25\% & 1/20 (5\%) \\
Diffusion Policy + Tactile & 14/20 (70\%) & 14/20 (70\%) & 6/20 (30\%) & 5/20 (25\%) & 48.75\% & 0/20 (0\%) \\
Finetuned Pi0.5 & 16/20 (80\%) & 16/20 (80\%) & 13/20 (65\%) & 6/20 (30\%) & 63.75\% & 2/20 (10\%) \\
\textbf{TacVLA (Ours)} & \textbf{20/20 (100\%)
}  & \textbf{18/20 (90\%)
} & \textbf{14/20 (70\%)
} & \textbf{15/20 (75\%)
} & \textbf{83.75\%}
 & \textbf{14/20 (70\%)
} \\
\bottomrule
\end{tabular}
\label{tab:overall}
\end{table*}

\subsection{Data Collection and Training Procedure}\label{sec:training}
To train and evaluate TacVLA, we collect a real-world dataset including four constraint-locked disassembly tasks and one in-box picking task. Each task contains 50 demonstrations collected through human teleoperation. The dataset consists of synchronized visual observations, language instructions, tactile maps, robot proprioception, and ground-truth action trajectories. Visual observations are captured from a fixed front camera and a wrist-mounted camera, while language instructions provide task-level prompts for each task, as listed in Table~\ref{tab:prompt_description}. All modality streams are recorded at 10Hz and temporally aligned to ensure synchronized observations at each timestep.

For each task, we fine-tune a separate task-specific model using the corresponding demonstration dataset. TacVLA is fine-tuned using Low-Rank Adaptation (LoRA) on top of an OpenPI backbone checkpoint (\texttt{pi05\_base}). All models are fine-tuned for 10,000 gradient steps under consistent optimization settings, and the tactile encoder is jointly optimized during fine-tuning to learn task-relevant tactile token representations.

\section{Experiments}
In this section, we evaluate the performance of our TacVLA model on a set of disassembly and a in-box picking task, then compare it against baseline models. We also conduct robustness evaluations to assess the model's performance under various conditions, such as visual occlusion and human disturbance.

\subsection{Experimental Setup}
Our experimental platform, shown in Fig.~\ref{fig:hardware}, consists of a 7-DoF Franka Emika Panda robotic arm equipped with parallel grippers. 
For visual perception, we employ two RGB cameras: a fixed front-facing camera providing a global scene view, and a wrist-mounted camera offering close-range observations.
To capture physical interaction signals, we integrate a tactile array sensor, adapted from~\cite{huang20253dvitac}, mounted on the surface of one of the fingers. 
The sensor provides a $15 \times 8$ tactile force distribution map at each timestep, corresponding to 120 normal contact force measurements over the sensing surface.
These tactile readings encode contact distribution across the sensor surface and are used as additional input to the policy during manipulation.

\begin{figure*}
    \centering
    \includegraphics[width=0.68\textwidth]{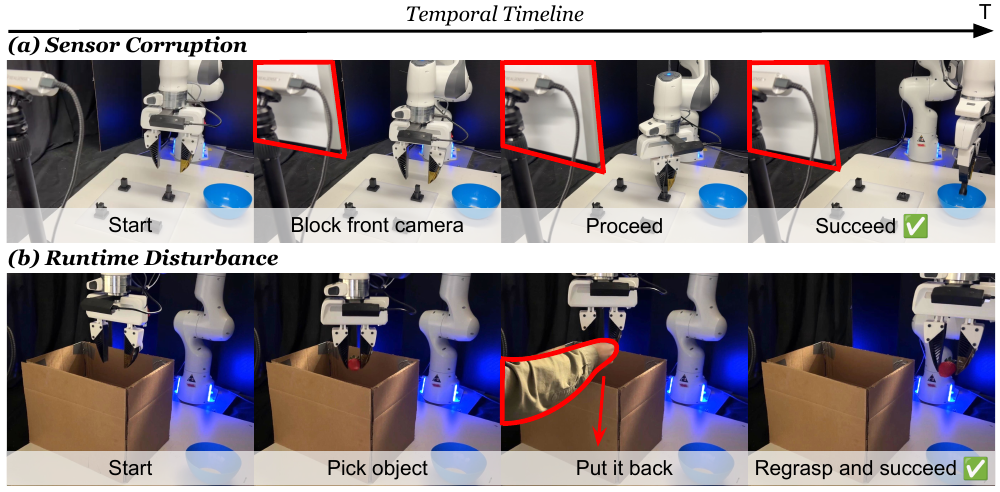}
    \caption{Robustness Evaluation: we evaluate the performance of our TacVLA model under conditions of visual occlusion and runtime disturbance to demonstrate its ability to adapt and maintain performance in challenging scenarios.}
    \label{fig:robustness_evaluation}
\end{figure*}

\begin{figure}
    \centerline{\includegraphics[width=\linewidth]{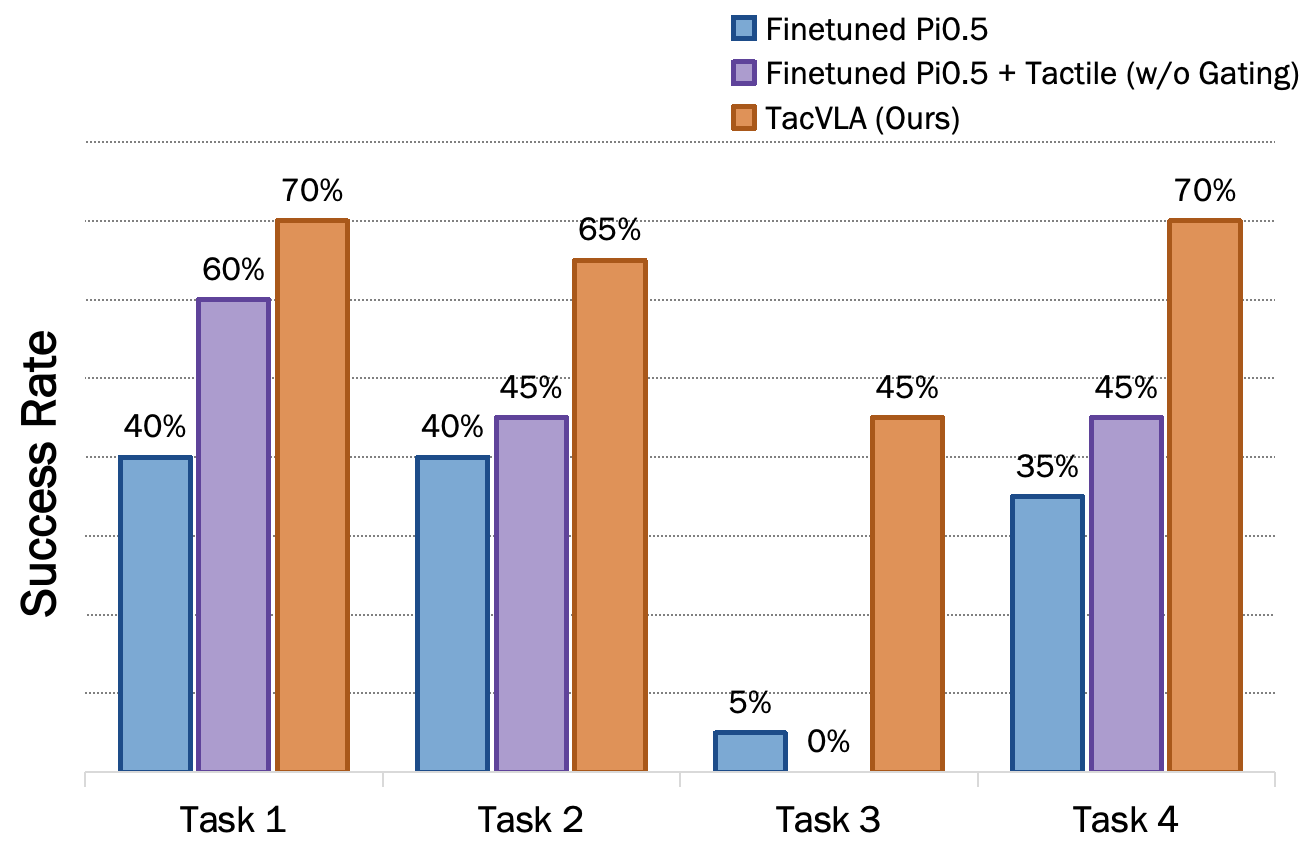}}
    \vspace{-0.6em}
    \caption{Success rates under block camera disturbance across four constraint-locked disassembly tasks. We compare finetuned Pi0.5, finetuned Pi0.5 with tactile input (w/o gating), and TacVLA (ours). Our method consistently improves performance under visual occlusion.}
    \label{fig:robustness_bar_chart}
    
\end{figure}

\subsubsection{Disassembly}
We design a set of contact-rich constraint-locked disassembly tasks (Fig.~\ref{fig:disassembly}) where we structure scenarios that require fine-grained adjustments based on tactile feedback. In these tasks, components are locked by different geometric constraints and must be separated through specific motions such as pressing, twisting, or sliding. 
The specific disassembly instructions for each task are illustrated in Fig.~\ref{fig:real_world_setup}.
These tasks require precise manipulation and the ability to adapt to changing contact conditions, making them ideal for evaluating the benefits of tactile sensing in a VLA framework. 

\subsubsection{In-box Picking}
We also conduct an in-box manipulation task (Fig.~\ref{fig:real_world_setup}), where the robot must retrieve objects from a confined space with limited visual access: (1) front camera has no access to box inside, (2) wrist camera has limited illumination and occlusion inside the box. 
This task further emphasizes the importance of tactile sensing for handling scenarios with severe visual occlusion and complex contact interactions, as the robot must rely heavily on tactile feedback to successfully retrieve the objects. 

\subsubsection{Benchmark with Diffusion Policy}

We further compare TacVLA with two diffusion-based baselines under comparable training settings: image-based Diffusion Policy~\cite{diffusion_policy} and 3D Diffusion Policy~\cite{DP3}. Both baselines take multi-view images and the same tactile measurements as input, encode each modality with standard visual backbones, and concatenate the resulting features to condition the diffusion model for action generation.
All policies are trained for 200 epochs, and the final checkpoints are used for evaluation.


\subsection{Results and Analysis}
Table.~\ref{tab:overall} reports the success rates across four disassembly tasks and in-box picking task among our model, Pi0.5 baseline and diffusion policy baseline. 
\subsubsection{Disassembly}
Our TacVLA achieves the best overall performance, with an average success rate of 83.75\%, significantly outperforming the finetuned Pi0.5 baseline (63.75\%) and diffusion-based policies.

While the Pi0.5 baseline performs reasonably well on Task 1 and Task 2 (both 80\%), its performance drops substantially on Task 3 (65\%) and Task 4 (30\%), where contact-dependent adjustments are critical.
In contrast, TacVLA maintains consistently strong performance across all tasks, achieving 100\% and 90\% success rates on Task 1 and Task 2, and improving to 70\% and 75\% on Task 3 and Task 4, respectively.

Notably, the largest gain is observed in Task 4 with 45\% improvement, where it involves fine-grained sliding interactions with partial occlusion.
In Task 4, we observe that the Pi0.5 baseline frequently becomes stuck in intermediate states and attempts repeated re-grasping, indicating uncertainty in contact dynamics.
In contrast, TacVLA demonstrates more stable behavior during contact transitions, suggesting that incorporating tactile sensing with state-dependent gating significantly enhances performance in fine-grained disassembly tasks.

\subsubsection{In-box Picking}
The in-box picking task presents severe visual challenges: the front camera has no visibility inside the box, and the wrist camera suffers from limited illumination and frequent occlusion.
Under these conditions, the finetuned Pi0.5 baseline achieves only 10\% success rate.
We observe that the baseline often attempts to execute the lifting motion without successfully grasping the object, indicating a lack of reliable in-hand state awareness in the absence of tactile feedback.

In contrast, TacVLA achieves a 70\% success rate, representing a substantial improvement under heavy occlusion.
Despite poor lighting and occasional object shifts due to imperfect initial grasps, TacVLA actively explores within the box and performs multiple re-grasp attempts.
Importantly, it proceeds to the lifting and placing stage only after contact is physically detected.

These results demonstrate that incorporating tactile sensing with contact-aware gating enables more reliable contact verification during exploration, significantly enhancing the robustness in scenarios dominated by visual occlusion and uncertain physical interaction.

\begin{table*}[!t]
\centering
\caption{Ablation study on the effect of the contact-aware gating module}
\renewcommand{\arraystretch}{1.5}
\setlength{\tabcolsep}{4pt}
\begin{tabular}{lcccccc}
\toprule
\multicolumn{1}{c}{\textbf{Method}} & \multicolumn{5}{c}{\textbf{Constraint-locked Disassembly}} & \textbf{In-Box Picking} \\
\cmidrule(lr){2-6}
 & \textbf{Task 1} & \textbf{Task 2} & \textbf{Task 3} & \textbf{Task 4} & \textbf{Average} &  \\
\midrule
Finetuned Pi0.5 + Tactile (w/o Gating)  & 18/20 (90\%) & 16/20 (80\%) & 12/20 (60\%) & 11/20 (55\%) & 71.25\% & 8/20 (40\%) \\
\textbf{TacVLA (Ours)} & \textbf{20/20 (100\%)
}  & \textbf{18/20 (90\%)
} & \textbf{14/20 (70\%)
} & \textbf{15/20 (75\%)
} & \textbf{83.75\%}
 & \textbf{14/20 (70\%)
} \\
\bottomrule
\end{tabular}
\label{tab:ablation}
\end{table*}

\subsubsection{Benchmark with Diffusion Policy}
We further compare TacVLA with diffusion-based policies augmented with tactile input.
On disassembly, the diffusion policy achieves an average success rate of 48.75\%, while the 3D diffusion policy achieves 31.25\%, both substantially lower than TacVLA's 83.75\%.
Under severe visual occlusion in the in-box picking task, diffusion policies perform poorly (0\% and 5\%, respectively).
Qualitatively, diffusion-based policies frequently struggle with accurate target localization, establishing stable contact, and maintaining consistent trajectories after contact.
In contrast, TacVLA exhibits more stable grasping and smoother contact transitions.

We attribute this difference primarily to the architectural and training paradigms. TacVLA builds upon a pretrained multimodal backbone and is fine-tuned via LoRA, enabling it to leverage rich visual-language priors and structured cross-modal representations. In contrast, diffusion policies are typically trained from scratch on task-specific datasets, which may limit their robustness and generalization under complex contact-rich conditions.


\subsection{Robustness Evaluation}
To evaluate the robustness of our TacVLA model, we conduct a series of evaluations under varying conditions that challenge the model's ability to perform contact-rich manipulation tasks.
\subsubsection{Block Camera}
As a critical test of the model's reliance on visual information,  we evaluate performance under severe visual occlusion by physically blocking the front camera during task execution, as shown in Fig.~\ref{fig:robustness_evaluation}(a). This setting simulates realistic failure scenarios such as occlusion, sensor corruption, or limited field of view, where reliable visual feedback is unavailable.

Fig.~\ref{fig:robustness_bar_chart} reports the success rates on four disassembly tasks under this disturbance. The vision-only baseline (finetuned Pi0.5) experiences a substantial performance drop, achieving 40\%, 40\%, 5\%, and 35\% success rates across Tasks 1–4, respectively. In contrast, TacVLA maintains significantly higher performance, reaching 70\%, 65\%, 45\%, and 70\%. Notably, Task 3 exhibits the largest improvement (+40\%), highlighting the model's ability to leverage tactile feedback when visual cues are severely degraded.
Overall, TacVLA improves the average success rate from approximately 30\% to over 60\% under visual occlusion, demonstrating enhanced robustness through state-dependent tactile integration.


\begin{figure}
    \centerline{\includegraphics[width=0.62\linewidth]{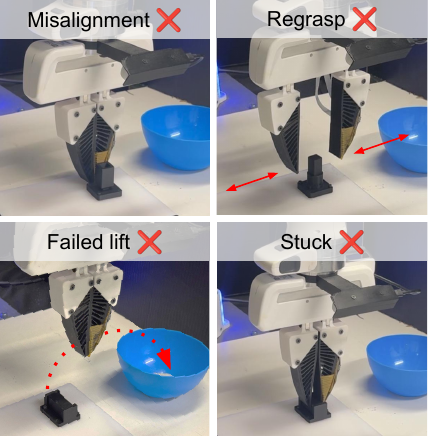}}
    \caption{Failure cases of \textit{Finetuned Pi0.5 + Tactile (w/o Gating)} during constraint-locked disassembly. Without contact-aware gating, tactile tokens remain active even in non-contact phases, leading to unstable behaviors.}
    \label{fig:failure_cases}
\end{figure}

\subsubsection{Human Disturbance}
To further assess robustness under dynamic environmental changes, we conduct a qualitative human-disturbance evaluation, as illustrated in Fig.~\ref{fig:robustness_evaluation}(b). During the in-box picking task, after the robot has grasped the object and begins lifting it out of the box, a human operator unexpectedly moves the object back into the box. This disturbance creates an out-of-distribution state change after partial task completion, and is designed to evaluate whether the policy can recognize the changed task state, adapt its subsequent actions, and recover from external perturbations.

TacVLA is able to recover from this disturbance: after the object is placed back into the box, the robot returns to the box, re-localizes and re-grasps the object, and then continues task execution. This behavior suggests that tactile feedback, together with contact-aware gating, helps the policy verify physical contact and adjust its behavior when the task state changes unexpectedly. In contrast, the finetuned Pi0.5 baseline fails to recover after the same disturbance and is unable to retrieve the object again, indicating limited adaptability under unexpected environmental perturbations.

\subsection{Ablation Study}
To isolate the effect of contact-aware gating, we compare TacVLA with a baseline that directly concatenates tactile tokens without applying the proposed contact-aware gating module, denoted as \textit{Finetuned Pi0.5 + Tactile (w/o Gating)}. Results across all tasks including robustness evaluations are reported in Table~\ref{tab:ablation} and Fig.~\ref{fig:robustness_bar_chart}.

\subsubsection{Disassembly}
On the disassembly tasks, removing gating reduces the average success rate from 83.75\% to 71.25\%.
Notably, on Task 3, the no-gating variant performs worse than the Pi0.5 baseline without tactile input (60\% vs 65\%), indicating that naive tactile fusion can even degrade performance.

Qualitatively, as shown in Fig.~\ref{fig:failure_cases}, the no-gating variant frequently exhibits misalignment during object approach, repeated re-grasp attempts, stalled intermediate states, and failed lift attempts.
These behaviors suggest that unconditional tactile token injection may interfere with visual localization before stable contact is established.


\subsubsection{In-box Picking Tasks}
On the in-box picking task, performance drops from 70\% to 40\% without gating.
Failure cases commonly involve repeated grasp attempts without successful completion.

This suggests that unconditional tactile fusion limits the model's ability to effectively leverage contact-state information under degraded visual conditions. In contrast, contact-aware gating enables a more stable transition from visual localization to physical interaction.

\subsubsection{Block Camera on Disassembly}
Under severe visual occlusion (front camera blocked), removing the gating mechanism results in a substantial performance drop, reducing the average success rate by approximately 25\% compared to TacVLA, as shown in Fig.~\ref{fig:robustness_bar_chart}.

This degradation indicates that contact-aware gating plays a critical role in leveraging tactile feedback when visual information is unreliable. Without gating, continuous participation in tactile tokens appears insufficient to compensate for visual degradation, highlighting the importance of state-dependent tactile activation for robust multimodal integration.

\section{Discussion and limitation}
Although TacVLA improves performance in contact-rich manipulation, several limitations remain. 
First, the contact-aware gating relies on a binary threshold heuristic, which does not allow gradual or learnable adjustment of modality importance. More adaptive modality weighting strategies could further improve cross-modal coordination.
Second, the tactile sensor provides efficient contact-state signals but has limited spatial resolution, restricting fine-grained contact geometry reasoning.
Finally, our evaluation focuses on short-horizon, contact-centric tasks; extending state-dependent modality routing to longer-horizon and more complex task settings remains future work.

\section{Conclusion}
We proposed TacVLA, a vision-language-action model augmented with tactile sensing for contact-rich manipulation. 
Through unified tokenization and contact-aware gating, TacVLA enables state-dependent multimodal fusion and robust cross-modal grounding. 
Experiments on disassembly, in-box picking task and robustness evaluations show clear improvements over pure VLA and diffusion-based policies.

\section*{Acknowledgment}
This work was supported in part by National Science Foundation under Award 2423068 and Award 2520136; and in part by the U.S. Department of Agriculture under Award 2023-67021-39072 and Award 2024-67021-42878.

\bibliographystyle{unsrt}
\bibliography{root}

\end{document}